\documentclass{article}

\usepackage{graphicx} 
\usepackage{subfigure} 

\usepackage{times}

\usepackage{natbib}

\usepackage{algorithm}
\usepackage{algorithmic}
\usepackage{float}

\usepackage{hyperref}
\usepackage{amssymb}
\usepackage{amsmath}

\usepackage[accepted]{arxiv} 

\icmltitlerunning{End-to-End Text Recognition with Hybrid HMM Maxout Models}

\DeclareMathOperator*{\argmax}{\arg\!\max}

\begin{document} 

\twocolumn[
\icmltitle{End-to-End Text Recognition with Hybrid HMM Maxout Models}

\icmlauthor{Ouais Alsharif}{ouais.alsharif@mail.mcgill.ca}
\icmladdress{Reasoning and Learning Laboratory, School of Computer Science, McGill University, Montreal, QC, Canada}
\icmlauthor{Joelle Pineau}{jpineau@cs.mcgill.ca}
\icmladdress{Reasoning and Learning Laboratory, School of Computer Science, McGill University, Montreal, QC, Canada}

\icmlkeywords{vision, deep learning, text recognition, hybrid model}

\vskip 0.3in
]

\begin{abstract} 

The problem of detecting and recognizing text in natural scenes has proved to be more challenging than its counterpart in documents, with most of the previous work focusing on a single part of the problem. In this work, we propose new solutions to the character and word recognition problems and then show how to combine these solutions in an end-to-end text-recognition system. We do so by leveraging the recently introduced Maxout networks along with hybrid HMM models that have proven useful for voice recognition. Using these elements, we build a tunable and highly accurate recognition system that beats state-of-the-art results on all the sub-problems for both the ICDAR 2003 and SVT benchmark datasets.\footnote{Code for this paper will be provided with the final version.}

\end{abstract} 

\section{Introduction}
Recognizing text in natural images is a challenging problem with many promising applications. Natural attributes such as lighting, shadows, styles, fonts and backgrounds that affect the perception of textual information is the main culprit that shifts this problem away from document text recognition, and closer to a combination of object and handwriting recognition. 

The end-to-end text recognition problem can be decomposed into three natural sub-problems: text detection, character recognition and word recognition. The first problem is to identify text locations in a natural image. The second problem requires identifying characters in cropped image patches; this is a classification problem with high confusion due to upper-case/lower-case and letter/number confusion. The third problem is a sequencing problem, given an image of a word, to output the most likely word corresponding to that image. These problems clearly overlap as character recognition is a sub-problem of word recognition, which itself is a sub-problem of text detection.

The end-to-end problem presents technical challenges on multiple levels. On the character level, the main challenge is to achieve high recognition accuracy. On the word level, the word recognizer needs to be accurate, fast, and scalable with lexicon size. On the end-to-end level, the system needs to balance precision, recall, complexity and speed. In our work, we aim for a highly accurate character recognizer, a fast, accurate and scalable word recognizer and for the end-to-end system, we aim for fast performance at query time, and high F-measure.

To achieve the goals specified above, we dissect the end-to-end recognition problem, exploring new solutions for all three subproblems. More specifically, we leverage a deep, convolutional variant of the recently introduced Maxout networks which make heavy use of dropout to beat the state-of-the-art on the character recognition task with minimal preprocessing. Also, inspired by the recent breakthroughs in voice recognition \cite{hinton2012deep}, we propose to treat the word recognition problem in a way similar to the phone-sequence recognition problem. More concretely, we create a hybrid HMM/Maxout architecture that is able to sequence words into their corresponding characters. The proposed model allows for simple integration of a lexicon's higher order n-grams, resulting in a method that is fast, accurate and highly tunable, while taking constant time relative to lexicon size. We then show how to integrate these parts in a novel end-to-end recognition system that achieves state-of-the-art F-measure on the ICDAR 2003 \cite{lucas2003icdar} and SVT \cite{wang2010word} datasets.

Due to the hierarchical dependency of the problem, where word recognition is a sub-module of text detection and character recognition is a sub-module of word recognition, we present the modules starting with character recognition in section 4, proceeding on to word recognition in section 5, and then joining the previous parts in the end-to-end system in section 6, with a pipeline of the end-to-end system in Figure 6. 

\section{Related Work}
The text recognition problem has been addressed in the literature on multiple levels: character recognition \cite{saidane2007automatic, coates2011text}, word recognition \cite{novikova2012large,mishra2012scene} and text detection \cite{hanif2008cascade,chen2011robust}. Most previous works focused on a single stage of the pipeline, with few looking into the end-to-end systems, namely \cite{wang2011end,neumann2011method,wang2012end}.

The character recognition problem is a classification problem that is generally addressed with the use of strong classifiers such as Convolutional Neural Nets (CNNs) \cite{wang2011end,wang2012end}, deformable parts models \cite{shiscene} or manually-engineered feature-extraction followed by a classifier \cite{decharacter}.

The word recognition problem is, much like phone recognition and handwriting recognition, a sequence recognition problem. Previous works have addressed this problem using CNNs \cite{wang2012end}, Conditional Random Fields (CRFs) \cite{mishra2012scene,novikova2012large} and Pictorial Structures (PS) \cite{wang2011end}. Most of the work in this area has relied on segmentation-free lexicon-dependent approaches. The use of the lexicons helps tackle the high confusion inherent in the text recognition problem. However, despite the argument for the validity of task-specific lexicon use in \cite{wang2011end}, it is clear that we ultimately wish to recognize text with a very general lexicon. To do so, we require word recognizers that scale well in the size of the lexicon. The works of \cite{neumann2011method,mishra2012scene,novikova2012large} are the only works we know of that show how their methods scale with lexicon size.

The text detection problem is defined such that, given a natural image, the goal is to output bounding boxes on all words in the image. Abstractly speaking, the problem is an instance of the object detection problem, followed by segmenting text regions into their constituent words. Previous works investigated different approaches for text detection, typically trading off precision, recall, training time and time consumed for manually designing features. Pre-trained CNNs \cite{coates2011text,wang2012end} applied in a multi-scale sliding window fashion are highly accurate but very time consuming. Viola-Jones style classifiers remedy the slowness in CNNs, but have long training times and require manually-engineered features \cite{hanif2008cascade,chen2011adaboost}. Alternative methods that cleverly exploit the nature of text such as Maximally Stable Extremal Regions (MSERs) \cite{matas2004robust} and Stroke Width Transform \cite{epshtein2010detecting} generally have lower accuracy but are fast to compute. Such methods were used successfully to detect text as in \cite{neumann2011method,chen2011robust,wang2011end}.

\section{Technical Background}
This section provides technical background on some of the techniques used in the proposed system.

\subsection{Convolutional Neural Networks}
Convolutional Neural Networks (CNNs) \cite{lecun1998gradient} are discriminativly trained neural networks that alternate convolution layers and pooling layers, with the last layer being usually a softmax or RBF layer. In a convolution layer, an input is convolved with multiple learned filters leading to multiple maps which then are pooled together through a pooling scheme. Combined with regularization and pretraining techniques, these neural nets achieve state-of-the-art results on many datasets \cite{krizhevsky2012imagenet,Goodfellow_Maxout_2013}.

\subsection{Dropout}
Dropout \cite{hinton2012improving} is a simple and efficient technique that can be used to reduce overfitting in neural networks. The main idea of dropout is to stochasticly omit some of the units from the network during learning. Intuitively, dropout adds robustness to the network by introducing noise on all levels of the architecture. Another way to view dropout is as a way to do model averaging over exponentially-many models with shared parameters. 

\subsection{Maxout Networks}
A Maxout network \cite{Goodfellow_Maxout_2013} is a multi-layer perceptron that makes heavy use of dropout to regularize the neural net, thereby reducing overfitting. It also uses a $max$ activation function that produces a sparse gradient. Specifically, in these networks, for an input $x \in \mathbb{R}^d $, every hidden layer implements the function:
$$h_i(x) = \max z_{ij},$$
where
$$z_{ij} = x^TW_{...ij}+b_{ij}$$
$$W\in \mathbb{R}^{d \times m \times k}, b\in \mathbb{R}^{m  \times k}.$$

Maxout networks have produced state-of-the-art results on benchmark datasets without any pre-training \cite{Goodfellow_Maxout_2013}.

\subsection{Hybrid HMM models}

HMMs have long been among the main tools used for sequence modelling in voice recognition \cite{rabiner1989tutorial} and hand-writing recognition \cite{hu1996HMM}. Hybrid models \cite{morgan1995continuous} extend HMMs with a simple idea, that is, instead of using Gaussian Mixture Models (GMMs) to model the HMMs observation model, hybrid models use Bayes rule and implicitly model the observation model using a probabilistic classifier. Concretely, let $O$ be a sequence of observations and let $Q$ be a state sequence, the purpose of the HMM is to produce $\displaystyle \argmax_Q p(Q|O)$. In a standard setting, to train an HMM, we require an observation model $p(o|q)$ where $o$ is an observation and $q$ is an HMM state. In the hybrid model, we approximate the observation model through Bayes rule with a probabilistic classifier that computes the posterior $p(q|o)$ distribution on HMM states $q$ given an input $o$. Concretely:
\begin{align}
p(o|q) \;\; = \;\; p(q|o)\frac{p(o)}{p(q)} \;\; \propto \;\; \frac{p(q|o)}{p(q)},
\end{align}

with $p(o)$ assumed to be equal for all observations.

Such hybrid models are usually trained with the embedded Viterbi algorithm \cite{bourlard1998hybrid} to maximize the likelihood of the data. In other variants of the model, hybrid models are discriminatively trained to optimize segmentation accuracy directly~\cite{bengio1992global,bengio1995lerec}.

Combined with deep architectures, these models have increased accuracies on challenging sequencing tasks primarily in voice-recognition \cite{hinton2012deep}.

\section{Character Recognition}
The character recognition problem involves building a character recognizer that when presented with a character image, produces a probability distribution over all characters. In our case, the recognizer classifies characters into 62 classes (26 upper-case, 26 lower-case and 10 digits). 

The dataset we use for this task is the ICDAR 2003 character recognition dataset \cite{lucas2003icdar} which consists of 6114 training samples and 5379 test samples after removing all non-alphanumeric characters as in \cite{wang2012end}. We augment the training dataset with 75,495 character images from the Chars74k English dataset \cite{decharacter} and 50,000 synthetic characters generated by \cite{wang2012end} making the total size of the training set 131,609 tightly cropped character images.

The architecture we use for this task is a five-layer convolutional Maxout network with the first three layers being convolution-pooling Maxout layers, the fourth a Maxout layer and finally a softmax layer on top. The first three layers have respectively 48, 128, 128 filters of sizes 8-by-8 for the first two and 5-by-5 for the third, pooling over regions of sizes 4-by-4, 4-by-4 and 2-by-2 respectively, with 2 linear pieces per Maxout unit and a 2-by-2 stride. The 4th layer has 400 units and 5 linear pieces per Maxout unit, fully connected with the softmax output layer.

We train the proposed network on 32-by-32 grey-scale character image patches with a simple preprocessing stage of subtracting the mean of every patch and dividing by its standard deviation + $\epsilon$. Similar to \cite{Goodfellow_Maxout_2013}, we train this network using stochastic gradient descent with momentum and dropout to maximize $\log p(y|x)$. Training was done on GPUs using Theano \cite{bergstra+al:2010-scipy} and pylearn \cite{pylearn2_arxiv_2013}.

The resulting character recognizer achieves state-of-the-art recognition rates on the ICDAR 2003 character test set with an accuracy of 85.5\% on the 62-way case-sensitive benchmark and 89.9\% on the case-insensitive 36-way benchmark. When we use the Maxout network as a feature extractor and feed the  features from the penultimate layer into an SVM with an RBF kernel, the recognition accuracy increases to 86\% on the 62-way benchmark while it remains roughly the same (89.8\%) on the 36-way benchmark. Table 1 compares our results to other works on this dataset.  Overall, the performance of the Maxout networks is slightly superior to that of the CNNs used in previous approaches.

As a side note, we found that different forms of binarization (otsu, random walkers, grabcuts) and preprocessing methods, such as ZCA as used in \cite{wang2012end}, do not enhance the test accuracy and in some cases,  decrease it.

\begin{table}[!ht]
\begin{center}
\caption{Character recognition accuracy on ICDAR 2003 test set. All methods use the same augmented training dataset.}
\vspace{1em}
\begin{tabular}{c|c|c}
    Work                & Method & Result \\

    \cite{coates2011text} & pre-trained CNNs & 81.7  \\
    \cite{wang2012end}    & pre-trained CNNs & 83.9  \\
    \textbf{this work}    & Conv-Maxout& \textbf{85.5}  \\
    \textbf{this work}    & Conv-Maxout + SVM& \textbf{86.0}  \\
\end{tabular}
\label{tab:char_recognition_results}
\end{center}
\end{table}
\vspace{-2em}

\section{Word Recognition}
The purpose of the word recognition module is to transcribe a word image into text. Our approach for word recognition is a segmentation-based, lexicon-free approach that could easily incorporate a lexicon during inference or as a post-inference processing tool. 

As it currently stands, it is difficult to recognize words with high accuracy without any language model due to the character confusion problem, therefore, all of the previous systems rely on lexicons to better the results. However, since lexicons can be very large, we make the distinction in our approach between where query time is linear in the size of the lexicon and those approaches where it is constant. 

To recognize a word, we first segment it into possible characters using a Hybrid HMM/Maxout model (Sec.5.1), then use the resulting segmentation to construct a cascade of potential characters (Sec.5.2), after which we apply a variant of the Viterbi algorithm that trades off speed and accuracy to compute a list of candidate results (Sec.5.3). Figure 1 depicts the pipeline for the full word recognition module.

\begin{figure}[!ht]
\centering
\includegraphics{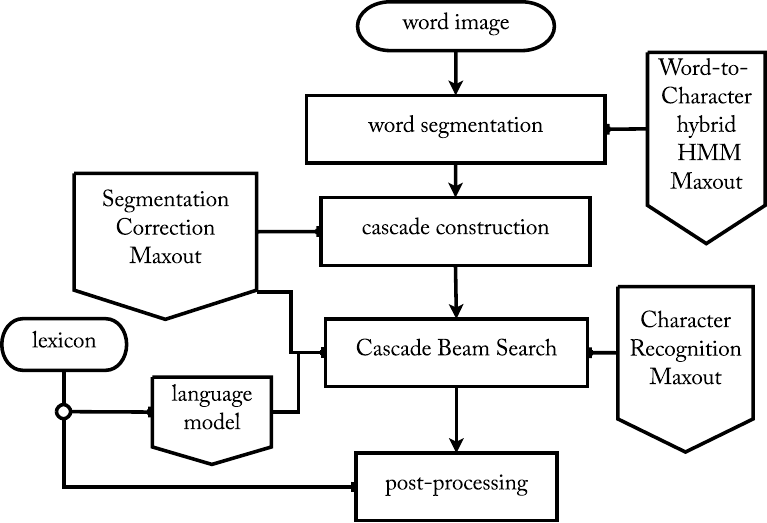}
\caption{Flowchart for the word recognition module. Pentagons represent learned modules. The character recognition Maxout module was described in Sec.4. Note that the lexicon can be used either as a post-processing tool, or during inference after a language model is constructed.}
\end{figure}

\subsection{Hybrid HMM Maxout Model}

The use of a hybrid HMM Maxout model for text segmentation was inspired by works in voice recognition \cite{renals1994connectionist,hinton2012deep}. Whereas in voice recognition the hybrid model is used to directly sequence phonemes, here we use it to segment word images into character/inter-character regions.

The dataset we use for training the model is made from the first 500 words in the ICDAR 2003 training-set. Specifically, for every word in the dataset, we create a segmentation into character/inter-character regions as follows: for every word, for every character pair that are adjacent, we define an inter-character region as the region stretching 10\% into the left character and 10\% into the right character.

In the hybrid model we use, depicted in Figure 2, the HMM has four states for each of the character/inter-character regions. For the classifier, we use a four-layer convolutional Maxout net with the first three layers being convolution/pooling layers with 48 filters each, where the filters were of size 8-by-8 for the first two layers and 5-by-5 for the third and a softmax layer on top. The first three layers had 2, 2 and 4 linear pieces per Maxout unit respectively and pooling was done on regions of size 4-by-4 with a 2-by-2 stride.

\begin{figure}[!ht]
\centering
\includegraphics{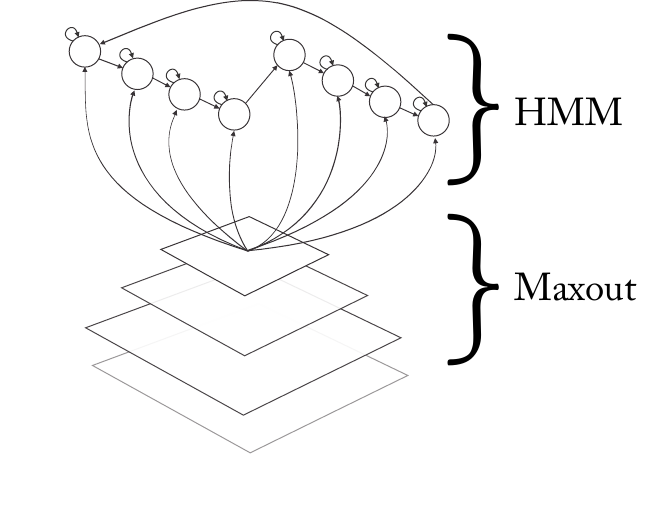}
\caption{Word-to-character Hybrid HMM Maxout module. The Maxout network's fourth layer is coupled with the HMM.}
\end{figure}

After creating the dataset, we train the Hybrid HMM model using embedded Viterbi training as defined in \cite{bourlard1998hybrid}. Somewhat surprisingly, we find that a single iteration of the embedded Viterbi is sufficient for the hybrid model to learn to segment; this is likely because the Maxout component is learning a good initial segmentation. After the model is trained, we use it to produce the segmentation $Q$ such that $\displaystyle \argmax_Q P(Q|O)$ with the standard Viterbi algorithm.

\subsection{Constructing the Cascade}

The segmentation produced by the hybrid model suffers from two natural shortcomings: over- and under-segmentation. Over-segmentation arises because some characters are composed by concatenating other characters, e.g. $V V$ instead of $W$. Under-segmentation is more often observed in cases of difficult fonts, blurry images and complex background noise.

To filter out instances of over-segmentation, we train a 4-layer convolutional Maxout network with the same architecture as the Maxout used in Sec.5.1, to predict the probability of over and under-segmentation. This network (called \textit{Segmentation Correction Maxout} in Fig.1) is trained on correct, over-, and under-segmentations generated from the ICDAR 2003 training dataset. We create a new interval from every two adjacent intervals if the joined interval has a higher probability of being a correct segmentation than both of its constituents under the learned network. As for under-segmentation, we simply divide every resulting interval into two intervals by cutting the resulting intervals in the middle. 

This operation produces what we call a cascade. Every cascade induces an adjacency graph that we use later for inferring the corresponding word. Figure 1 depicts a cascade for the word ``JFC" where the middle row is the segmentation from the hybrid HMM/Maxout model along with its induced graph.

\begin{figure}[ht!]
\centering
\includegraphics{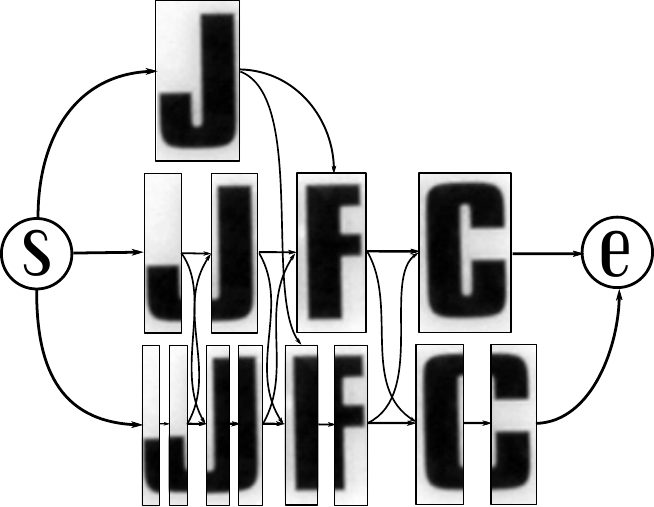}
\caption{A sample cascade with its induced graph. The middle row is the output of the Hybrid HMM Maxout module. The letter J in the top level was produced as an output of the Segmentation Correction Maxout module. The lower row is produced by systematic splitting of all the nodes in the middle row.}
\end{figure}

\subsection{Word Inference}

Computing the most likely word given a cascade is equivalent to computing the most likely path from the beginning of the cascade to its end. This problem can be solved using dynamic programming in a way similar to the Viterbi algorithm, except where both nodes and edges in the graph incur a cost.

Let the alphabet consist of $K$ characters, also, let $c_i$ be character with index $i$, let $v_k$ be an interval in the cascade with index $k$ where the cascade intervals are sorted by their left-most point. We define $S(c_i,v_k)$ to be the probability of the most likely sequence ending in interval $v_k$ and character $c_i$. We also define $N(v_k)$ to be the set of intervals that immediately precede $v_k$.
$S$ can be computed optimally using a Viterbi style algorithm in two cases: with no language model, or with a bigram language model. In the first case $S$ becomes:
\begin{equation}
S(c_i,s_k) = p(c_i|s_k)p(s_k) \max_{j,q} S(c_j,s_q),
\end{equation}
while in the second:
\begin{equation}
S(c_i,s_k) = p(c_i|s_k)p(s_k) \max_{j,q} p(c_i|c_j)S(c_j,s_q),
\end{equation}
such that $q \in N(v_k)$.

Computing $S$ takes $O(K^2V + V \log(V))$ where $V$ is the number of intervals in the cascade assuming that for any $v_k$, $|N(v_k)| = O(1)$. The most likely word can be found by tracing back the largest $S(c_i,v_k)$ for all intervals whose end is the end of the cascade\footnote{One issue with the optimization problem in equation 3 is that it is comparing sequences on different probability spaces such that the sequence length prior is induced by the hybrid model's segmentation and the cascade construction. This particular issue was not directly studied by any works that we know of. Other works in the area have handled it in other ways, for references, consult \cite{bengio1995lerec,lecun1998gradient}}.

While we can obtain $p(c|v)$ as the posterior from the five-layer character recognition Maxout network (from section 4), we obtain $p(v)$ from the Segmentation Correction Maxout network specified in Sec.5.2. As for the language model, $p(c_i|c_j)$, we obtain it from a predefined lexicon. 

\begin{table*}[!th]
\centering
\caption{Word recognition accuracies on ICDAR 2003 and SVT datasets}
\vspace{1em}
\begin{tabular}{c|c|c|c|c|c}

Work & Method & W-Small & W-Medium & W-large & SVT\\
\cite{wang2012end}  & CNNs & 90.0 & 84.0 &- & 70.0 \\ 
\cite{mishra2012scene} & CRFs & 81.8 & 67.8 &- & 73.2\\
\cite{novikova2012large} & CRFs & 82.8 & - &- & 72.9\\
\cite{wang2011end} & PS & 76.0 & 62.0 &- & 57.0 \\ 
This work, 5-gram language model & HMM/Maxout & 90.1& 87.3 & 83.0 & 67.0\\
\textbf{This work, edit-distance} & HMM/Maxout & \textbf{93.1} & \textbf{88.6} & \textbf{85.1} & \textbf{74.3} \\
\end{tabular}\\
\end{table*}

The straight forward generalization of equation (3) to n-gram language models incurs a large time penalty on the order of $O(K^n)$. To side step that penalty while allowing for higher order language models we propose an algorithm that trades off accuracy with inference time in a way similar to Beam Search \cite{russell1995artificial}; keeping the top $B$ candidates for every interval. We call this Cascade Beam Search (see Algorithm 1). Here, $p(c|lm,w)$ is the probability of a character given an n-gram language model $lm$ and a sequence of characters $w$ that come before it, $cost_v$ is the visual cost of an interval character pair, $cost_l$ is the linguistic cost of ending in an interval $s$ with a character $c$, and $\parallel$ is the string concatenation operation. 

\begin{algorithm}[th!]
   \caption{Cascade Beam Search}
   
   \label{alg:example}
\begin{algorithmic}
   \STATE {\bfseries Input:} intervals $s_i$, language model $lm$	, Beam width $B$\\
   \FOR{$i=1$ {\bfseries to} $V$}
   \FOR{$j \in N(v_i)$}
   \FOR{$c_k \in Alphabet$}
   \FOR{every word $w$ in $Q_j$}
   \STATE $\hat{w}$ = $w \parallel c_k$
   \STATE $cost_v = p(c_k|v_i)*p(v_i)$
   \STATE $cost_l = p(c_k|lm,w)$
   \STATE $cost_{\hat{w}} = cost_v * cost_l *cost_w$
   \STATE Add $(\hat{w},cost_{\hat{w}})$ to $Q_i$ 
   \IF{ $size(Q_i) > B$}
   \STATE remove word with lowest cost from $Q_i$
   \ENDIF
   \ENDFOR
   \ENDFOR
   \ENDFOR
   \ENDFOR
   \STATE {\bfseries return} all $w \in Q_j$ sorted decreasingly by their costs, such that $v_j$ is at the end of the word
\end{algorithmic}
\end{algorithm}

\subsection{Word Recognition Results}
We test our word recognition subsystem on word images from the ICDAR 2003 \cite{lucas2003icdar} and SVT \cite{wang2010word} word recognition test sets. The ICDAR 2003 test set consists of images of tightly cropped words. The SVT test set is a harder benchmark with more loosely cropped words and case-wise incorrect labellings. Similar to \cite{wang2011end,wang2012end}, all of our tests are on words that do not contain non-alphanumeric characters and that are of length greater than 2, leaving 860 and 647 test words for the ICDAR 2003 and SVT datasets respectively.

For the ICDAR 2003 test set, we test the recognizer under three scenarios that vary by lexicon size: small, medium and large. In the case of small lexicons, an image's lexicon contains the ground truth word in the image in addition to 50 distractor words provided by \cite{wang2011end}. In the medium lexicon case, the lexicon contains all the words in the test set. For the large lexicon case, we use Hunspell's spell checking dictionary\footnote{available here: http://wordlist.sourceforge.net/} that contains almost 50,000 words, in addition to all the words in the test set. We call these scenarios respectively W-Small, W-Medium and W-Large.  

As for the SVT test set, as in \cite{wang2011end,wang2012end}, we test the recognizer under a single setting in which for every word, we use other distractor words provided in the dataset. Moreover, since the SVT dataset's lexions contain only capitalized words, we collapse the classifier's result $p(c_k|v_i)$ by setting the probability of a character to the sum of its upper-case and lower-case probabilities.

We also test the system in two modes, the first is where the system takes constant time in lexicon size per query, while in the second we permit the query to post-process the result with a lexicon.

In the first mode, we test the system with language models constructed from task-specific lexicons. While in the second mode, we do not use a language model at all and instead, we use the resulting list of words from the cascade beam search algorithm and consider the recognition result to be the most likely resultant word that exists in the lexicon, or in case none of the resulting words were in the lexicon, we use the word with the least edit distance to any word in the lexicon. 

Table 2 compares our results on the benchmarks W-Small, W-Medium, W-Large and SVT to previous published results under the two modes specified above. All experiments were run with a beam width $B=100$. Without the use of either a language model or a lexicon, the module reaches an accuracy of 55.6\%.

As is shown in table 2, our proposed algorithm outperforms previous state-of-the-art algorithms on the specified benchmarks. On the large lexicon benchmark, we couldn't find works that were directly comparable to ours. However, we note that when we increase the lexicon size a 1000-fold, we get an accuracy of 85.1\% which compares favourably with 78\% achieved by \cite{novikova2012large} when they increase their lexicon size 90-fold.

\subsubsection{Effect of Beam Width}

Since the complexity of the cascade beam search algorithm is $O(KVB \log(B) + V \log(V) )$, we could trade the accuracy of the algorithm with its speed through the parameter $B$. Figure 4 shows the effect of the beam width on recognition accuracy and on recognition speed on the W-Small task. As shown in the figure, a small beam width does not lead to a great decrease in accuracy, and permits a great increase in recognition speed, making the word recognition module almost 15 times faster.

\begin{figure}[!ht]
\centering
\includegraphics{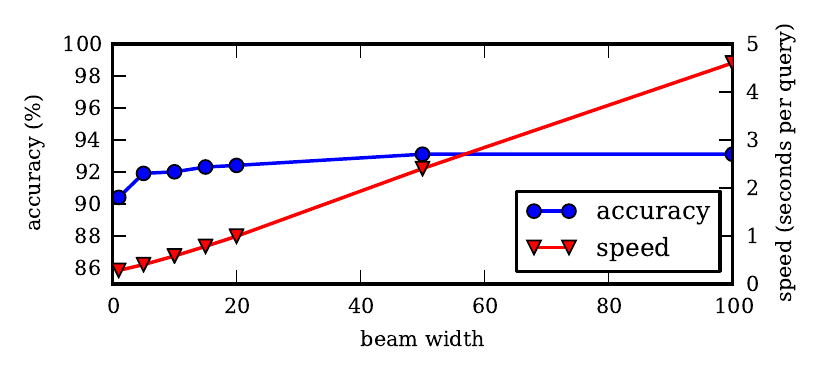}
\caption{Beam Width vs. Accuracy on the ICDAR 2003 word recognition dataset under a small lexicon scenario}
\end{figure}

\subsubsection{Effect of Language Model Order}
As noted earlier, the Cascade Beam Search algorithm also allows for integration of higher order language models directly through the inference stage. This would be most helpful in the cases of very large lexicons, since the inference process takes a constant time in lexicon size after the initial stage of encoding the lexicon by its n-grams. Figure 5 depicts how accuracy changes with the language model's order for different lexicon sizes. The Small, Medium and Large curves correspond to using the Small, Medium and Large lexicons specified in section 5.4. The Large* curve corresponds to using the same large lexicon but without adding the ground truth words in the lexicon; this is the only scenario done on case-insensitive words. The highest accuracy reached under the Large* scenario is 67.0\%. It is notable that in the Large* scenario, higher orders of language models cause overfitting and thereby reduce the recognition accuracy.

\begin{figure}[!ht]
\centering
\includegraphics{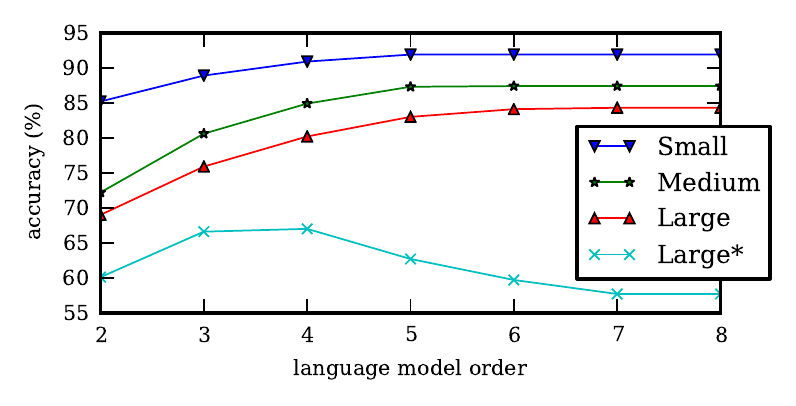}
\caption{Language model order vs. accuracy by lexicon size on the ICDAR 2003 test set with beam width $B=100$. Note that the Small, Medium and Large curves are tested on case-sensitive words while the large* is on case-insensitive words}
\end{figure}
\begin{table*}[!ht]
\centering
\caption{End-to-end F-measures on the ICDAR 2003 and SVT datasets}
\vspace{1em}

\begin{tabular}{c|c|c|c|c|c|c}
Work & I-5& I-20& I-50 & I-Full & I-Large & SVT\\
\cite{wang2011end} & 72 & 70 & 68 & 51 & - & 38\\
\cite{wang2012end} & 76 & 74 & 72 & 67 & - & 46\\
\textbf{This work} & \textbf{80} & \textbf{79} & \textbf{77} & \textbf{70} & \textbf{63} & \textbf{48} \\
\end{tabular}\\
\end{table*}
\vspace{-1em}
\section{End-to-End Text Recognition}
In this section, we show how the previous parts can be integrated into a full end-to-end text recognition system. The main issue, unaddressed by the previous sections, is to localize text patches in natural images.

\subsection{End-to-End Pipeline}

To extract text locations from an image, we start by getting possible text candidates using Maximally Stable Extremal Regions (MSERs). MSERs are defined to be regions in the image that are either maximas or minimas of image intensities with respect to their surroundings. While being highly imprecise text detectors, they can be computed very quickly \cite{nister2008linear}. The use of MSERs allows us to sidestep the enormous time penalty incurred by applying a costly recognizer on multiple scales of the image as in \cite{wang2012end}, thereby allowing our system to become much more efficient. Since MSERs would ideally correspond to character regions, we form candidate line boxes by clustering the character candidates with DBSCAN \cite{ester1996density} using multiple distances to obtain candidate line-level bounding boxes.

After we obtain the line-level bounding boxes, we segment these lines using the \textit{Line-to-Word Hybrid HMM/Maxout} trained to segment lines to words from the ICDAR 2003 scene training set. We then add words by gradually introducing gaps from the segmentation one at a time in descending size order. After this, we threshold the resulting word bounding boxes using the \textit{Word Detection Maxout}, a four-layer convolutional Maxout network with the same architecture as the one used in Sec.5.1 on word/non-text images extracted from ICDAR 2003 scene training dataset. We also threshold words on the $cost_v$ score resulting from the word recognition module and the edit-distance value.

\begin{figure}[!ht]
\centering
\includegraphics{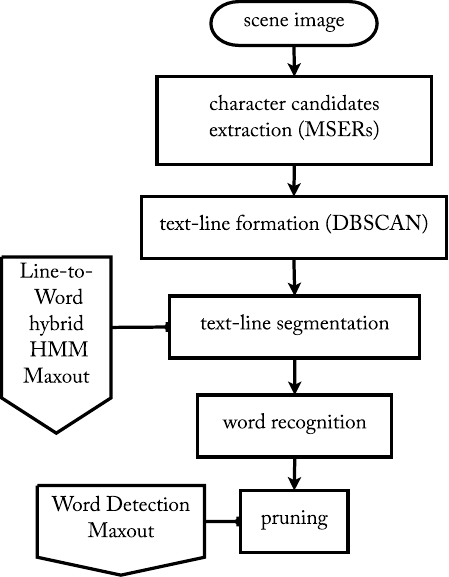}
\caption{End-to-end pipeline. Pentagons represent learned modules. The word recognition module shown here represents the full system from Fig.1.}
\end{figure}

We follow this pipeline by doing a non-max suppression (NMS) \cite{neubeck2006efficient} on word boxes that overlap by 30\% of the area of their bounding box according to the visual cost $cost_v$ from the word recognition module.

\subsection{End-to-End Results}
We tested the above system on both the ICDAR 2003 and SVT end-to-end scene text-recognition test sets. Each of the datasets contain 249 scene images. More specifically, for the ICDAR 2003 dataset, we conduct tests under five scenarios, where for the first three, the lexicons consist if \{5,20,50\} distractor words per image in addition to the ground truth words for that image, in the fourth scenario all the test words are included in the lexicon and in the fifth scenario, we use the same large lexicon we used to test the word recognition module (Sec.5.4). We label these scenarios I-5, I-20, I-50, I-Full and I-Large respectively. The lexicons were provided by the authors of \cite{wang2011end}. As for the SVT dataset, we conduct the tests using the lexicons provided with the dataset. All tests were done with the text-recognition module in the edit-distance mode.

\begin{figure*}[!ht]
\centering
\includegraphics{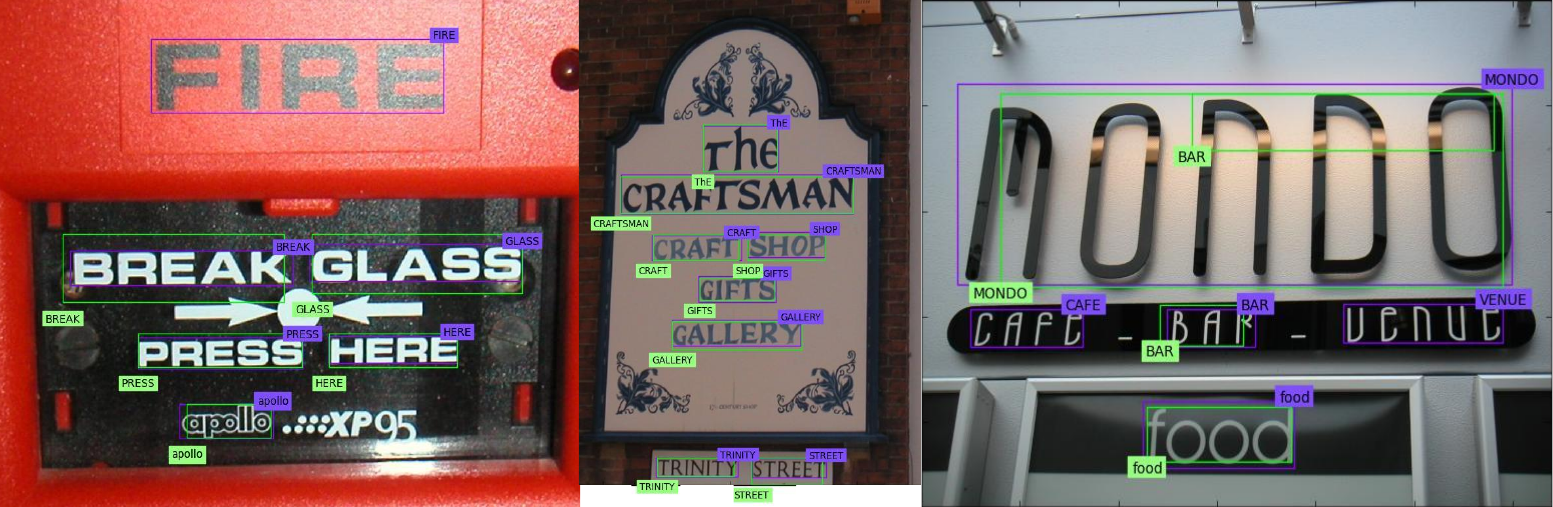}
\caption{Samples from the end-to-end results, the purple boxes represent the ground truth and the green boxes represent the predictions}
\end{figure*}

We test the end-to-end system using the standard precision/recall metrics under the benchmarks specified in \cite{lucas2003icdar}, where a prediction is considered a hit when the area of the overlap between the predicted box and the target box is greater than 50\% of the bounding box area and the predicted text matches exactly. 

Table 3 compares our results to other results in the field.  Despite our use of a simple method with low accuracy like MSERs to extract possible text regions, our end-to-end system is able to outperform previous state-of-the-art end-to-end systems and produce reasonable results for large lexicons. Figure 8 shows the precision/recall curves for all the tasks on the ICDAR 2003 dataset and Figure 7 shows a few sample outputs from our system.

\begin{figure}[!ht]
\centering
\includegraphics{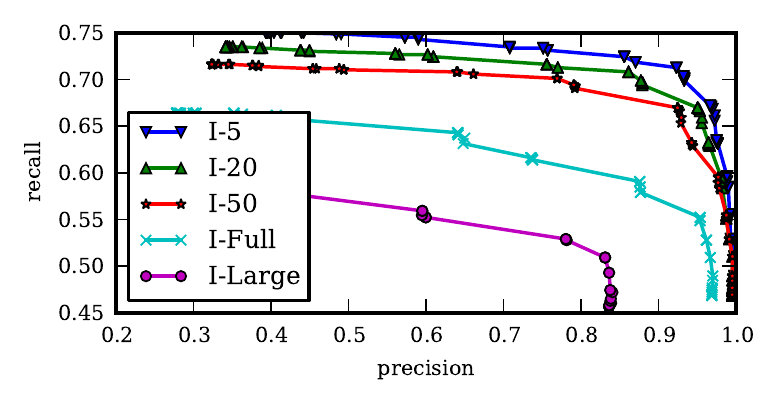}
\caption{Precision/recall curves for the end-to-end system on the ICDAR 2003 dataset under different lexicon sizes}
\end{figure}

\section{Discussion}
In this section, we discuss possible ways to increase the accuracy of both the word recognition module and the end-to-end system, and possible ways to make the entire system operate under real-time constraints.

For the word recognition module, using learned edit-distances \cite{ristad1998learning,mccallum2012conditional} would help boost the module's accuracy. Beyond that, most of the loss in accuracy comes from segmentations created by the hybrid HMM model. Designing a neural net to factor in context information into the hybrid HMM while computing posterior probabilities should help reduce that loss in accuracy.

To increase the F-measures on the end-to-end system, we should seek to boost recall. As pointed out in \cite{neumann2011method} MSERs do not offer high recall for character location extraction. The alternative of using a time-consuming but highly accurate classifier as in \cite{wang2012end} is not practical to make the end-to-end system work in real-time. In our opinion, a promising solution would be to develop a Viola-Jones-style cascade \cite{viola2001rapid} coupled with feature-learning. Such an approach could offer a fast, accurate, easy to train and feature-engineering-free text detector that would increase recall. 

\section{Conclusion}

In this work, we presented a novel end-to-end text recognition system. We proposed novel solutions to each subproblem in the end-to-end system. Specifically, we leveraged convolutional Maxout networks to beat the state-of-the-art on character recognition. We showed how to use the character recognizer in a word recognizer that is fast, tunable, highly accurate, and scales elegantly with lexicon size. We then constructed an end-to-end text recognition system using the previous modules in addition to other, relatively simple constructs. The proposed system outperforms previous works on end-to-end text recognition tasks on standard challenging benchmarks.

\section{Acknowledgements}

We would like to thank Yoshua Bengio, Aaron Courville and Ian Goodfellow for their insightful comments and discussions. We would also like to thank Paul Kry and Sheldon Andrews for providing the GPUs. Members of the RL lab for helpful discussions. Financial support for this research was provided by the NSERC Discovery grant.

\pagebreak
{\small
\pagebreak

\bibliographystyle{arxiv_copy}
\bibliography{arxiv_copy}

\begin{thebibliography}{37}
\providecommand{\natexlab}[1]{#1}
\providecommand{\url}[1]{\texttt{#1}}
\expandafter\ifx\csname urlstyle\endcsname\relax
  \providecommand{\doi}[1]{doi: #1}\else
  \providecommand{\doi}{doi: \begingroup \urlstyle{rm}\Url}\fi

\bibitem[Bengio et~al.(1992)Bengio, De~Mori, Flammia, and
  Kompe]{bengio1992global}
Bengio, Y, De~Mori, R, Flammia, G, and Kompe, R.
\newblock Global optimization of a neural network-hidden markov model hybrid.
\newblock \emph{Neural Networks, IEEE Transactions on}, 3\penalty0
  (2):\penalty0 252--259, 1992.

\bibitem[Bengio et~al.(1995)Bengio, LeCun, Nohl, and Burges]{bengio1995lerec}
Bengio, Y, LeCun, Y, Nohl, C, and Burges, C.
\newblock Lerec: A nn/hmm hybrid for on-line handwriting recognition.
\newblock \emph{Neural Computation}, 7\penalty0 (6):\penalty0 1289--1303, 1995.

\bibitem[Bergstra et~al.(2010)Bergstra, Breuleux, Bastien, Lamblin, Pascanu,
  Desjardins, Turian, Warde-Farley, and Bengio]{bergstra+al:2010-scipy}
Bergstra, J, Breuleux, O, Bastien, F, Lamblin, P, Pascanu, R, Desjardins, G,
  Turian, J, Warde-Farley, D, and Bengio, Y.
\newblock Theano: a {CPU} and {GPU} math expression compiler.
\newblock In \emph{Proceedings of the Python for Scientific Computing
  Conference ({SciPy})}, 2010.

\bibitem[Bourlard \& Morgan(1998)Bourlard and Morgan]{bourlard1998hybrid}
Bourlard, H and Morgan, N.
\newblock Hybrid hmm/ann systems for speech recognition: Overview and new
  research directions.
\newblock In \emph{Adaptive Processing of Sequences and Data Structures}. 1998.

\bibitem[Chen et~al.(2011)Chen, Tsai, Schroth, Chen, Grzeszczuk, and
  Girod]{chen2011robust}
Chen, H, Tsai, S~S, Schroth, G, Chen, D~M, Grzeszczuk, R, and Girod, B.
\newblock Robust text detection in natural images with edge-enhanced maximally
  stable extremal regions.
\newblock In \emph{ICIP}, 2011.

\bibitem[Chen \& Yuille(2011)Chen and Yuille]{chen2011adaboost}
Chen, X and Yuille, A.
\newblock Adaboost learning for detecting and reading text in city scenes.
\newblock 2011.

\bibitem[Coates et~al.(2011)Coates, Carpenter, Case, Satheesh, Suresh, Wang,
  Wu, and Ng]{coates2011text}
Coates, A, Carpenter, B, Case, C, Satheesh, S, Suresh, B, Wang, T, Wu, D~J, and
  Ng, A~Y.
\newblock Text detection and character recognition in scene images with
  unsupervised feature learning.
\newblock In \emph{ICDAR}, 2011.

\bibitem[de~Campos et~al.(2009)de~Campos, Babu, and Varma]{decharacter}
de~Campos, T~E, Babu, B~R, and Varma, M.
\newblock Character recognition in natural images.
\newblock 2009.

\bibitem[Epshtein et~al.(2010)Epshtein, Ofek, and
  Wexler]{epshtein2010detecting}
Epshtein, B, Ofek, E, and Wexler, Y.
\newblock Detecting text in natural scenes with stroke width transform.
\newblock In \emph{CVPR}, 2010.

\bibitem[Ester et~al.(1996)Ester, Kriegel, Sander, and Xu]{ester1996density}
Ester, M, Kriegel, H, Sander, J, and Xu, X.
\newblock A density-based algorithm for discovering clusters in large spatial
  databases with noise.
\newblock 1996.

\bibitem[Goodfellow et~al.(2013{\natexlab{a}})Goodfellow, Warde-Farley,
  Lamblin, Dumoulin, Mirza, Pascanu, Bergstra, Bastien, and
  Bengio]{pylearn2_arxiv_2013}
Goodfellow, I, Warde-Farley, D, Lamblin, P, Dumoulin, V, Mirza, M, Pascanu, R,
  Bergstra, J, Bastien, F, and Bengio, Y.
\newblock Pylearn2: a machine learning research library.
\newblock \emph{arXiv:1308.421}, 2013{\natexlab{a}}.

\bibitem[Goodfellow et~al.(2013{\natexlab{b}})Goodfellow, Warde-Farley, Mirza,
  Courville, and Bengio]{Goodfellow_Maxout_2013}
Goodfellow, I~J., Warde-Farley, D, Mirza, M, Courville, A, and Bengio, Y.
\newblock Maxout networks.
\newblock In \emph{ICML}, 2013{\natexlab{b}}.

\bibitem[Hanif et~al.(2008)Hanif, Prevost, and Negri]{hanif2008cascade}
Hanif, S, Prevost, L, and Negri, P.
\newblock A cascade detector for text detection in natural scene images.
\newblock In \emph{ICPR}, 2008.

\bibitem[Hinton et~al.(2012{\natexlab{a}})Hinton, Deng, Yu, Dahl, Mohamed,
  Jaitly, Senior, Vanhoucke, Nguyen, Sainath, et~al.]{hinton2012deep}
Hinton, G, Deng, L, Yu, D, Dahl, G~E, Mohamed, A, Jaitly, N, Senior, A,
  Vanhoucke, V, Nguyen, P, Sainath, Tara~N, et~al.
\newblock Deep neural networks for acoustic modeling in speech recognition: The
  shared views of four research groups.
\newblock \emph{Signal Processing Magazine, IEEE}, 29\penalty0 (6):\penalty0
  82--97, 2012{\natexlab{a}}.

\bibitem[Hinton et~al.(2012{\natexlab{b}})Hinton, Srivastava, Krizhevsky,
  Sutskever, and Salakhutdinov]{hinton2012improving}
Hinton, G~E, Srivastava, N, Krizhevsky, A, Sutskever, I, and Salakhutdinov,
  R~R.
\newblock Improving neural networks by preventing co-adaptation of feature
  detectors.
\newblock \emph{arXiv:1207.0580}, 2012{\natexlab{b}}.

\bibitem[Hu et~al.(1996)Hu, Brown, and Turin]{hu1996HMM}
Hu, J, Brown, M~K, and Turin, W.
\newblock Hmm based online handwriting recognition.
\newblock \emph{Pattern Analysis and Machine Intelligence, IEEE Transactions
  on}, 18\penalty0 (10):\penalty0 1039--1045, 1996.

\bibitem[Krizhevsky et~al.(2012)Krizhevsky, Sutskever, and
  Hinton]{krizhevsky2012imagenet}
Krizhevsky, A, Sutskever, I, and Hinton, G.
\newblock Imagenet classification with deep convolutional neural networks.
\newblock In \emph{NIPS}, 2012.

\bibitem[LeCun et~al.(1998)LeCun, Bottou, Bengio, and
  Haffner]{lecun1998gradient}
LeCun, Y, Bottou, L, Bengio, Y, and Haffner, P.
\newblock Gradient-based learning applied to document recognition.
\newblock \emph{Proceedings of the IEEE}, 86\penalty0 (11):\penalty0
  2278--2324, 1998.

\bibitem[Lucas et~al.(2003)Lucas, Panaretos, Sosa, Tang, Wong, and
  Young]{lucas2003icdar}
Lucas, S, Panaretos, A, Sosa, L, Tang, A, Wong, S, and Young, R.
\newblock Icdar 2003 robust reading competitions.
\newblock 2003.

\bibitem[Matas et~al.(2004)Matas, Chum, Urban, and Pajdla]{matas2004robust}
Matas, J, Chum, O, Urban, M, and Pajdla, T.
\newblock Robust wide-baseline stereo from maximally stable extremal regions.
\newblock \emph{Image and vision computing}, 22\penalty0 (10):\penalty0
  761--767, 2004.

\bibitem[McCallum et~al.(2012)McCallum, Bellare, and
  Pereira]{mccallum2012conditional}
McCallum, A, Bellare, K, and Pereira, F.
\newblock A conditional random field for discriminatively-trained finite-state
  string edit distance.
\newblock \emph{arXiv:1207.1406}, 2012.

\bibitem[Mishra et~al.(2012)Mishra, Alahari, Jawahar, et~al.]{mishra2012scene}
Mishra, A, Alahari, Karteek, Jawahar, CV, et~al.
\newblock Scene text recognition using higher order language priors.
\newblock 2012.

\bibitem[Morgan \& Bourlard(1995)Morgan and Bourlard]{morgan1995continuous}
Morgan, N and Bourlard, H.
\newblock Continuous speech recognition.
\newblock \emph{Signal Processing Magazine, IEEE}, 12\penalty0 (3):\penalty0
  24--42, 1995.

\bibitem[Neubeck \& Van~Gool(2006)Neubeck and Van~Gool]{neubeck2006efficient}
Neubeck, A and Van~Gool, L.
\newblock Efficient non-maximum suppression.
\newblock In \emph{ICPR}, 2006.

\bibitem[Neumann \& Matas(2011)Neumann and Matas]{neumann2011method}
Neumann, L and Matas, J.
\newblock A method for text localization and recognition in real-world images.
\newblock In \emph{ACCV}. 2011.

\bibitem[Nist{\'e}r \& Stew{\'e}nius(2008)Nist{\'e}r and
  Stew{\'e}nius]{nister2008linear}
Nist{\'e}r, D and Stew{\'e}nius, H.
\newblock Linear time maximally stable extremal regions.
\newblock In \emph{ECCV}. 2008.

\bibitem[Novikova et~al.(2012)Novikova, B, K, and L]{novikova2012large}
Novikova, T, B, Olga, K, Pushmeet, and L, Victor.
\newblock Large-lexicon attribute-consistent text recognition in natural
  images.
\newblock In \emph{ECCV 2012}. 2012.

\bibitem[Rabiner(1989)]{rabiner1989tutorial}
Rabiner, L.
\newblock A tutorial on hidden markov models and selected applications in
  speech recognition.
\newblock \emph{Proceedings of the IEEE}, 77\penalty0 (2):\penalty0 257--286,
  1989.

\bibitem[Renals et~al.(1994)Renals, Morgan, Bourlard, Cohen, and
  Franco]{renals1994connectionist}
Renals, S, Morgan, N, Bourlard, H, Cohen, M, and Franco, H.
\newblock Connectionist probability estimators in hmm speech recognition.
\newblock \emph{Speech and Audio Processing, IEEE Transactions on}, 2\penalty0
  (1):\penalty0 161--174, 1994.

\bibitem[Ristad \& Yianilos(1998)Ristad and Yianilos]{ristad1998learning}
Ristad, E and Yianilos, P.
\newblock Learning string-edit distance.
\newblock \emph{Pattern Analysis and Machine Intelligence, IEEE Transactions
  on}, 20\penalty0 (5):\penalty0 522--532, 1998.

\bibitem[Russell et~al.(1995)Russell, Norvig, Canny, Malik, and
  Edwards]{russell1995artificial}
Russell, S~J, Norvig, P, Canny, J~F, Malik, J~M, and Edwards, D~D.
\newblock \emph{Artificial intelligence: a modern approach}, volume~74.
\newblock Prentice hall Englewood Cliffs, 1995.

\bibitem[Saidane \& Garcia(2007)Saidane and Garcia]{saidane2007automatic}
Saidane, Z and Garcia, C.
\newblock Automatic scene text recognition using a convolutional neural
  network.
\newblock 2007.

\bibitem[Shi et~al.(2013)Shi, Wang, Xiao, Zhang, Gao, and Zhang]{shiscene}
Shi, C, Wang, C, Xiao, B, Zhang, Y, Gao, S, and Zhang, Z.
\newblock Scene text recognition using part-based tree-structured character
  detection.
\newblock 2013.

\bibitem[Viola \& Jones(2001)Viola and Jones]{viola2001rapid}
Viola, P and Jones, M.
\newblock Rapid object detection using a boosted cascade of simple features.
\newblock In \emph{Computer Vision and Pattern Recognition, 2001. CVPR 2001.
  Proceedings of the 2001 IEEE Computer Society Conference on}, volume~1, pp.\
  I--511. IEEE, 2001.

\bibitem[Wang \& Belongie(2010)Wang and Belongie]{wang2010word}
Wang, K and Belongie, S.
\newblock Word spotting in the wild.
\newblock In \emph{ECCV}. 2010.

\bibitem[Wang et~al.(2011)Wang, Babenko, and Belongie]{wang2011end}
Wang, K, Babenko, B, and Belongie, S.
\newblock End-to-end scene text recognition.
\newblock In \emph{ICCV}, 2011.

\bibitem[Wang et~al.(2012)Wang, Wu, Coates, and Ng]{wang2012end}
Wang, T, Wu, D, Coates, A, and Ng, A.
\newblock End-to-end text recognition with convolutional neural networks.
\newblock In \emph{ICPR}, 2012.

\end{thebibliography}
}

\end{document}